\documentclass[letterpaper, 10 pt, conference]{ieeeconf}  

\IEEEoverridecommandlockouts                              

\usepackage{graphicx}
\usepackage{caption}
\graphicspath{
    {figures/}
}

\usepackage{amsmath,amssymb,enumerate}

\usepackage{amsthm}
\usepackage{setspace}
\usepackage{booktabs}
\usepackage[usenames,dvipsnames,svgnames,table]{xcolor}
\usepackage{mathtools}
\usepackage{algorithm, algorithmicx, algpseudocode}
\usepackage{blindtext}
\usepackage{gensymb}
\usepackage{xparse}
\usepackage{lipsum}
\usepackage{mathrsfs}
\usepackage[mathscr]{euscript}
\usepackage{times}
\usepackage{multicol}
\definecolor{darkgreen}{rgb}{0,0.6,0}
\usepackage[bookmarks=true,colorlinks=true,pdfpagemode=UseNone,citecolor=darkgreen,linkcolor=black,urlcolor=BrickRed,pagebackref]{hyperref}
\usepackage[caption=false,font=footnotesize]{subfig}
\usepackage{amsfonts}
\usepackage[utf8]{inputenc}
\usepackage[T1]{fontenc}
\usepackage{textcomp}
\usepackage{arydshln}
\usepackage{balance}

\newtheorem{assumption}{Assumption}

\renewcommand{\thefigure}{\arabic{figure}}

\definecolor{note}{rgb}{0.1,0.1,1}
\definecolor{rephase}{rgb}{0.15,0.7,0.15}
\definecolor{bag}{rgb}{0.6,0.6,0.2}

\makeatletter
\renewcommand*\env@matrix[1][c]{\hskip -\arraycolsep
  \let\@ifnextchar\new@ifnextchar
  \array{*\c@MaxMatrixCols #1}}
\makeatother

\newcommand{\transpose}{\mathsf{T}}

\makeatletter
\newcommand{\mathleft}{\@fleqntrue\@mathmargin0pt}
\newcommand{\mathcenter}{\@fleqnfalse}
\makeatother

\definecolor{orange}{RGB}{255,127,0}

\title{\LARGE \bf Input Influence Matrix Design for MIMO Discrete-Time Ultra-Local~Model} 

\author{Sangli Teng, Amit K. Sanyal, Ram Vasudevan, Anthony Bloch, Maani Ghaffari%
\thanks{Toyota Research Institute provided funds to support this work. Funding for M. Ghaffari was in part
provided by NSF Award No. 2118818. A. Bloch was supported in part by NSF grant DMS-2103026 and AFOSR grant FA0550-18-0028.}
\thanks{S. Teng, R. Vasudevan, A. Bloch, and M. Ghaffari are with the University of Michigan, Ann Arbor, MI 48109, USA. \texttt{\{sanglit,ramv,abloch,maanigj\}@umich.edu}.}
\thanks{A. K. Sanyal is with the Department of Mechanical and Aerospace Engineering, Syracuse University, Syracuse, NY, USA. \texttt{aksanyal@syr.edu}.}
}

\begin{document}

\maketitle
\thispagestyle{empty}
\pagestyle{empty}

\begin{abstract}

Ultra-Local Models (ULM) have been applied to perform model-free control of nonlinear systems with unknown or partially known dynamics. Unfortunately, extending these methods to MIMO systems requires designing a dense input influence matrix which is challenging. This paper presents guidelines for designing an input influence matrix for discrete-time, control-affine MIMO systems using an ULM-based controller. This paper analyzes the case that uses ULM and a model-free control scheme: the Hölder-continuous Finite-Time Stable (FTS) control. By comparing the ULM with the actual system dynamics, the paper describes how to extract the input-dependent part from the lumped ULM dynamics and finds that the tracking and state estimation error are coupled. The stability of the ULM-FTS error dynamics is affected by the eigenvalues of the difference (defined by matrix multiplication) between the actual and designed input influence matrix. Finally, the paper shows that a wide range of input influence matrix designs can keep the ULM-FTS error dynamics (at least locally) asymptotically stable. A numerical simulation is included to verify the result. The analysis can also be extended to other ULM-based controllers.

\end{abstract} 

\IEEEpeerreviewmaketitle

\section{INTRODUCTION}
{An Ultra-Local Model (ULM) is a control affine model designed to locally represent a controlled dynamical system with unknown or partly known dynamics. When the input influence matrix is designed, the ULM dynamics can be estimated by model-free observers and applied to perform feedback control~\cite{sanyal2021discrete}. However, for a system with $n$ inputs and $m$ outputs, the input influence matrix has $n\times m$ entries in general, which is non-trivial to determine when its size becomes large.} To provide guidelines for the design of the input influence matrix in the MIMO ULM framework, we analyze the stability of the error dynamics considering the coupling effect of controller and observer. As the current digital controllers are implemented in discrete-time, we consider the case using the model-free Hölder-continuous Finite-Time Stable (FTS) control and estimation framework~\cite{sanyal2021discrete}. 

Intelligent PID (iPID)~\cite{ulm, iPID} control has been applied for model-free control using the ULM, where a lumped dynamics term represents the system dynamics locally. The iPID controller has been applied to the SISO system and requires the user to design the input coefficient via trial and error. For a MIMO system, the coefficient becomes a matrix (possibly dense), which makes applying the ULM framework difficult. Therefore, many applications assume the system can be decoupled into several SISO systems; thus, classical ULM-based iPID framework can be applied~\cite{ULM-siso1, ulm-siso2, ULM-MPC}. However, this assumption may not always be reasonable. Additionally, the trial and error method in parameter tuning has no guarantee of stability. The work of \cite{ULM-LMI} imposes  the Linear Matrix Inequality condition to obtain the optimal ULM gain. However, \cite{ULM-LMI} only considers linear time-invariant systems and the real input influence matrix is known. 

By using a Hölder-continuous Lyapunov function \cite{fts-lyap}, a FTS tracking controller has been proposed and applied in~\cite{cfts1, fts2, fts3}. An FTS state estimator has also been applied for unmanned aerial vehicle state estimation \cite{fts-obs}. In \cite{sanyal2021discrete}, the FTS control and estimation scheme is developed for the ULM framework, which enables finite-time stable learning and control of the unknown nonlinear system. Although most applications of ULM are in continuous time, the discrete-time system naturally considers time delay and is more advantageous in embedded systems. In our previous work \cite{sanyal2021discrete}, the discrete-time FTS tracking and state estimation scheme has been applied in the ULM framework. Similar to other applications of ULM, it is also a challenge to design the input influence matrix.

The main contribution of this work is the quantitative analysis of the stability of the ULM framework on the control affine system considering the design of the input influence matrix. As the designed input influence matrix in the ULM framework is different (almost always) from the real one, this discrepancy makes the ULM dynamics input-dependent. Therefore, the control input can affect the observer for the ULM dynamics, which makes the observer and controller coupled. Unlike the work of \cite{iPID} that assumes lumped dynamics terms, we extract the input-dependent part from the ULM dynamics. Based on this formulation, we derive the error dynamics that can guide the design of the input influence matrix. Furthermore, we show that the stability of the error dynamics is affected by the eigenvalues of the difference (defined by matrix multiplication) between the real and designed input influence matrix. 

The remainder of this paper is organized as follows. In Section \ref{prelim}, we provide the basic knowledge of the nonlinear discrete-time dynamics system, ultra-local model framework, and the Hölder-continuous finite-time stable control and state estimation scheme. Section \ref{stability} provides stability analysis of the coupled observer and controller error dynamics. Numerical analysis is provided in Section \ref{simresult} to present and verify the conditions under which the error dynamics are asymptotically stable. Section \ref{sec:discuss} discusses the limitations and future works. Finally, Section \ref{sec:conclusion} concludes the paper.




\section{Preliminaries}
\label{prelim}
This section introduces the basic concepts of a discrete-time nonlinear system, an ultra-local model of the unknown dynamics, and a Hölder-continuous finite-time stable state estimator and controller.

\subsection{Discrete-time nonlinear system}
Consider a nonlinear system with $m$ inputs, $n$ outputs and $l$ unknown parameters. The notation $(\cdot)_k = (\cdot)(t_k)$ denotes the value of time-varying variables at the time step $t_k$. We define $u_k \in \mathbb{R}^{m}$ to be the input vectors, $y_k \in \mathbb{R}^{n}$ to be outputs and $z_k \in \mathbb{R}^{l}$ to be unknown parameters. We define $k\in \mathbb{W} = \{0, 1, 2, ...\}$ and $\mathbb{W}$ to be the set of the whole numbers including $0$. 

We use the superscript $(\mu)$ to denote the $\mu$th order finite difference of a variable in discrete time: 
\begin{equation}
\label{eq:defdifference}
    y^{(\mu)}_k:=y^{(\mu-1)}_{k+1} - y^{(\mu-1)}_{k},\ \ y^{(0)}_k = y_k.
\end{equation}
Thus the unknown discrete time system can be expressed as: 
\begin{equation}
\label{eq:nonlineardiff}
y_{k}^{(v)}=\varpi\left(y_{k}, y_{k}^{(1)}, \ldots, y_{k}^{(v-1)}, z_{k}, u_{k}, t_{k}\right).
\end{equation}\\
Using \eqref{eq:defdifference}, we can convert the nonlinear system \eqref{eq:nonlineardiff} to:
\begin{equation}
\label{eq:nonlinearSys}
y_{k+v}=\varphi\left(y_{k}, y_{k+1}, \ldots, y_{k+v-1}, z_{k}, u_{k}, t_{k}\right) .
\end{equation}
We assume that the system \eqref{eq:nonlinearSys} can be represented in the control affine form:
\begin{equation}
\label{eq:system}
    y_{k+v} = F_{k} + G_{k}u_{k},
\end{equation}
where $F_k \in \mathbb{R}^{n}, G_k \in \mathbb{R}^{n\times m}, u_k \in \mathbb{R}^{m}$.
Note that not all dynamical systems are in the control affine form. But for a wide range of applications on robotics, this assumption is applicable. We denote the vectors of variables on which the system \eqref{eq:nonlinearSys} depends, as:  
\begin{equation}
    \chi_k := (y_k, y_{k+1}, y_{k+v-1}, z_k, u_k, t_k). 
\end{equation}   

We assume that the system has the following properties: 
\begin{assumption}\label{as:1}
  $F_{k}$ and $G_{k}$ are Lipschitz continuous with respect to the $\chi_k$.
\end{assumption}
\begin{assumption}\label{as:2}
  The numbers of inputs and outputs are the same, i.e., $m=n$. $G_{k}$ is invertible. 
\end{assumption}
These assumptions guarantee that the system \eqref{eq:system} is input-output controllable. 
\subsection{Ultra-local model of discrete-time nonlinear system}
To control the system without  knowledge of system dynamics $G_k$ and $F_k$, the ultra-local model (ULM) represents the system \eqref{eq:system} by:
\begin{equation}
\label{eq:ulm_system}
    y_{k+v} = \mathcal{F}_{k} + \mathcal{G}_{k}u_{k},
\end{equation}
where $\mathcal{F}_k \in \mathbb{R}^{n}, \mathcal{G}_k \in \mathbb{R}^{n\times m}$. $\mathcal{G}_k$ is unknown but can be identified or designed.  $\mathcal{F}_k$ is also unknown but can be obtained via state observers when $\mathcal{G}_k$ is determined. {It is worth noticing that the ULM model we use is only local and not unique even for a control-affine system, as the input influence matrix is obtained by design.}


{
\subsection{Hölder-continuous feedback}
We briefly introduce the discrete-time Hölder-continuous Lyapunov function \cite{sanyal2021discrete} as a prerequisite to the controller and observer design.  
We say that a discrete-time Lyapunov function $V_k:\mathbb{R}^n\rightarrow \mathbb{R}$ is Hölder-continuous if
\begin{equation}
    \label{eq:FTS-lyap}
    V_{k+1} - V_{k} \le -\gamma_k V_k^\alpha,
\end{equation}
where $\gamma_k$ is a positive definite function of $V_k$ satisfying the condition that $\exists \epsilon \in \mathbb{R}^{+}$:
$$ \gamma_k = \gamma(V_k) \ge \eta := \epsilon ^{1-\alpha}, \ V_k\ge \epsilon.$$
Lemma 2.1 and Theorem 2.1 in \cite{sanyal2021discrete} suggest that the $V_{k}$ converges to $0$ in finite steps if it satisfies \eqref{eq:FTS-lyap}. We will show the finite-time stability of the controller and observer by constructing the discrete-time Hölder-continuous Lyapunov function.
} 
\subsection{Model-free finite fime stable observer}
A first-order observer has been proposed in \cite{sanyal2021discrete} to estimate unknown dynamics $\mathcal{F}_k$ in the ULM \eqref{eq:ulm_system}. Let the estimation of $\mathcal{F}_k$ be $\hat{\mathcal{F}}_k$ and thus the estimation error becomes:
\begin{equation*}
\label{eq:erF}
    {e}^{\mathcal{F}}_k := \hat{\mathcal{F}}_k - \mathcal{F}_k,
\end{equation*}
The first order finite difference of $\mathcal{F}_k$ can be defined as:
\begin{equation*}
    \Delta \mathcal{F}_k:=\mathcal{F}_{k+1} - \mathcal{F}_k. 
\end{equation*}
According to Theorem 3.1 of \cite{sanyal2021discrete}, with $\gamma > 0$ and $r \in (1, 2)$, 
a first-order observer can be designed as: 
\begin{equation}
\label{FTSObserver}
    \begin{aligned}
\hat{\mathcal{F}}_{k+1} &=\mathcal{D}\left(e_{k}^{\mathcal{F}}\right) e_{k}^{\mathcal{F}}+\mathcal{F}_{k} \text { given } \hat{\mathcal{F}}_{0}, \\
\text { where } \mathcal{D}\left(e_{k}^{\mathcal{F}}\right) &=\frac{\left(\left(e_{k}^{\mathcal{F}}\right)^{\mathrm{T}} e_{k}^{\mathcal{F}}\right)^{1-1 / r}-\gamma}{\left(\left(e_{k}^{\mathcal{F}}\right)^{\mathrm{T}} e_{k}^{\mathcal{F}}\right)^{1-1 / r}+\gamma}.
\end{aligned}
\end{equation}
{
By Theorem 3.1 in \cite{sanyal2021discrete}, if the estimation error dynamics $e_{k}^{\mathcal{F}}$ satisfies $e^{\mathcal{F}}_{k+1}=\mathcal{D}\left(e_{k}^{\mathcal{F}}\right)e_{k}^{\mathcal{F}}$, $e_{k}^{\mathcal{F}}$ will converge to zero in finite time. The convergence property can be verified by defining the Lyapunov function ~\cite{sanyal2021discrete, fts-lyap}: $$V_k^{\mathcal{F}}:=(e_k^{\mathcal{F}})^{\transpose}e_k^{\mathcal{F}}.$$ 
Thus, we can find $\gamma_k$ and show it satisfies \eqref{eq:FTS-lyap},
\begin{equation*}
    \begin{aligned}
    V_{k+1}^{\mathcal{F}} - V_k^{\mathcal{F}} &= - \gamma^{\mathcal{F}}_{k} (V_k^{\mathcal{F}})^{\frac{1}{r}},\\
    \gamma^{\mathcal{F}}_{k}&:=(1-\mathcal{D}(e_k^{\mathcal{F}})^2)(V_k^{\mathcal{F}})^{1-\frac{1}{r}}.
    \end{aligned}
\end{equation*}
}
As $\mathcal{F}_{k+1}$ is not available at time step $k$, we substitute $\mathcal{F}_{k+1}$ with $\mathcal{F}_{k}$ thus introducing the perturbation term $\Delta \mathcal{F}_k$. The estimation error $e^{\mathcal{F}}_k$ is guaranteed to converge to a bounded neighborhood of zero as long as $\Delta \mathcal{F}_k$ is bounded ~\cite{sanyal2021discrete}.

\subsection{Model-free finite time stable controller}
Similar to the first-order observer, a finite-time stable tracking controller can also be designed based on the estimation of $\mathcal{F}_k$.  Theorem 4.1 of \cite{sanyal2021discrete} suggests that, with positive number $\mu > 0$ and $s \in (1,2)$, we can define the control law as:

\begin{equation}
\label{FTScontroller}
\begin{aligned}
\mathcal{G}_{k} u_{k} &=y_{k+v}^{d}-\hat{\mathcal{F}}_{k}+\mathcal{C}\left(e_{k+v-1}^{y}\right) e_{k+v-1}^{y}, \\
\mathcal{C}\left(e_{j}^{y}\right) &=\frac{\left(\left(e_{j}^{y}\right)^{\mathrm{T}} e_{j}^{y}\right)^{1-1 / s}-\mu}{\left(\left(e_{j}^{y}\right)^{\mathrm{T}} e_{j}^{y}\right)^{1-1 / s}+\mu}.
\end{aligned}
\end{equation}
{Similar to the observer design, we can define the Hölder-continuous Lyapunov function and show the finite-time stability for this control law.} The tracking error is expected to converge to a bounded neighborhood of zero as long as the ${e}^{\mathcal{F}}_k$ is bounded. 

\section{Stability Analysis of the Error Dynamics}
\label{stability}
In many applications with ULM, the controller and state estimator are designed separately. In this section shows that the ULM dynamics term $\mathcal{F}_k$ is input-dependent, thus making the tracking error and state estimation error coupled. Based on the coupled error dynamics, we could derive the conditions under which the errors are asymptotically stable. 

\subsection{Splitting the ultra-local model dynamics}
Considering the real system dynamics \eqref{eq:system} and its ULM representation \eqref{eq:ulm_system}, we define the difference between the real input influence matrix and the designed one: 
\begin{equation*}
    \Delta \mathcal{G}_k := G_k - \mathcal{G}_k.
\end{equation*}
Now the system dynamics \eqref{eq:system} can be written as:
\begin{equation*}
    y_{k+v} = (F_k + \Delta \mathcal{G}_k u_k) + \mathcal{G}_k u_k.
\end{equation*}
Comparing with \eqref{eq:ulm_system}, we find that the ULM dynamics also depends on the inputs:
\begin{equation}
    \label{eq:fhat-f}
    \mathcal{F}_k = F_k + \Delta\mathcal{G}_k u_k.
\end{equation}
Now we have split the ULM dynamics into the input-dependent part $\Delta\mathcal{G}_k u_k$ and the real system dynamics $F_k$. We can see that $\mathcal{F}_k$ is a function of state input $u_k$ that comes from the feedback controller. Later we will show the term $\Delta\mathcal{G}_k u_k$ makes the tracking and state estimation error coupled. 

\subsection{Dynamics of state estimation error}
Considering the observer \eqref{FTSObserver} and substituting the expression \eqref{eq:fhat-f}, we have the dynamics of $e^{\mathcal{F}}_k$:
\begin{equation}
\label{eFDyn}
\begin{aligned}
e^{\mathcal{F}}_{k+1}&=\hat{\mathcal{F}}_{k+1}-\mathcal{F}_{k+1} \\
&=\mathcal{D}\left(e_{k}^{\mathcal{F}}\right) e_{k}^{\mathcal{F}}+\mathcal{F}_{k}-\mathcal{F}_{k+1} \\
&=\mathcal{D}\left(e_{k}^{\mathcal{F}}\right) e_{k}^{\mathcal{F}}+(F_k+\Delta \mathcal{G}_{k}u_k)-(F_{k+1}+\Delta \mathcal{G}_{k+1}u_{k+1})\\
&=\mathcal{D}\left(e_{k}^{\mathcal{F}}\right) e_{k}^{\mathcal{F}}-\Delta F_{k} + (\Delta\mathcal{G}_{k}u_k - \Delta\mathcal{G}_{k+1}u_{k+1}),
\end{aligned}
\end{equation}
where $\Delta F_k := F_{k+1} - F_{k}$ is assumed to be bounded according to the Lipschitz continuity of $F_k$.  
Comparing to the error dynamics in \cite{sanyal2021discrete}, we find the dynamics is perturbed by the term $(\Delta\mathcal{G}_{k}u_k - \Delta\mathcal{G}_{k+1}u_{k+1})$. If the actual input influence matrix is known, such that $\Delta \mathcal{G}_k = 0$, the system dynamics is identical to that of \cite{sanyal2021discrete}, thus, having the same error convergence.

\subsection{Dynamics of tracking error}
To analyze the tracking error dynamics, we first express input $u_k$ by substituting \eqref{eq:fhat-f} and \eqref{eq:erF} into \eqref{FTScontroller}:
\begin{equation*}
\label{eq:ukderive}
\begin{aligned}
     u_k&=\mathcal{G}_{k}^{-1}(y_{k+v} - \hat{\mathcal{F}}_k + \mathcal{C}\left(e_{k+v-1}^{y}\right) e_{k+v-1}^{y}) \\
    &=\mathcal{G}_{k}^{-1}(y_{k+v} - (F_k + \Delta\mathcal{G}_k u_k + e^{\mathcal{F}}_k)  \\
    & \qquad \qquad \qquad \qquad + \mathcal{C}\left(e_{k+v-1}^{y}\right) e_{k+v-1}^{y}).
\end{aligned}
\end{equation*}
Extracting $u_k$, we have:
\begin{equation}
\label{eq:uk}
\begin{aligned}
  {{u}_{k}}&={{(I\text{+}\mathcal{G}_{k}^{-1}\Delta {{\mathcal{G}}_{k}})}^{-1}}\mathcal{G}_{k}^{-1}\delta_k \\
  &=({{\mathcal{G}_{k}^{-1}(\mathcal{G}_{k}+\Delta {{\mathcal{G}}_{k}})}})^{-1} \mathcal{G}_{k}^{-1}\delta_k \\
  &={{(\mathcal{G}_{k}+\Delta {{\mathcal{G}}_{k}})^{-1}}}\mathcal{G}_{k}\mathcal{G}_{k}^{-1}\delta_k = G^{-1}_k\delta_k \\
\delta_k &:= y_{k+v}^{d}-e_{k}^{\mathcal{F}}-{{{F}}_{k}}+\mathcal{C}\left(e_{k+v-1}^{y}\right) e_{k+v-1}^{y}.
\end{aligned}
\end{equation}
Substituting \eqref{eq:uk} into the system dynamics \eqref{eq:system}, we have:
\begin{equation}
\begin{aligned}
  & {{y}_{k+v}}={{F}_{k}}+{{G}_{k}}{{u}_{k}} = {{F}_{k}}+\delta_k \\
 & = {{F}_{k}} +  y_{k+v}^{d}-e_{k}^{\mathcal{F}}-{{{F}}_{k}}+\mathcal{C}\left(e_{k+v-1}^{y}\right) e_{k+v-1}^{y}.
\end{aligned}
\end{equation}
By reorganising the last equations, we can obtain the tracking error dynamics:
\begin{equation}
\label{eyDyn}
    e_{k+v}^{y}+e_{k}^{\mathcal{F}}=\mathcal{C}\left(e_{k+v-1}^{y}\right) e_{k+v-1}^{y}, 
\end{equation}
which is identical to that of \cite{sanyal2021discrete}. However, we also note that $e_{k+v}^{y}$ will appear in \eqref{eFDyn} when we substitute \eqref{eq:uk}.

\subsection{Coupled error dynamics}
With \eqref{eFDyn}, \eqref{eyDyn} and control input \eqref{eq:uk}, we have a discrete error dynamics that depends on $\Delta \mathcal{G}_k$:
\begin{equation}
\label{ErrorDynClose}
\begin{aligned}
  & e_{k+v}^{y}=\mathcal{C}(e_{k+v-1}^{y})e_{k+v-1}^{y}-e_{k}^{\mathcal{F}} \\ 
 & e_{k+v+1}^{y}=\mathcal{C}(e_{k+v}^{y})e_{k+v}^{y}-e_{k}^{\mathcal{F}} \\ 
 & e_{k+1}^{\mathcal{F}}=\mathcal{D}(e_{k}^{\mathcal{F}})e_{k}^{\mathcal{F}}-\Delta {{F}_{k}}+(\Delta {{\mathcal{G}}_{k}}{{u}_{k}}-\Delta {{\mathcal{G}}_{k+1}}{{u}_{k+1}}).
\end{aligned}
\end{equation}
The perturbation term $\Delta \mathcal{G}_ku_k$ can be converted to:
\begin{equation}
\label{Perturb}
\begin{aligned}
  & \Delta {{\mathcal{G}}_{k}}{{u}_{k}}=\Delta {{\mathcal{G}}_{k}}G_{k}^{-1}(y_{k+v}^{d}-e_{k}^{\mathcal{F}}-{{\mathcal{F}}_{k}}+Cee) \\ 
 & =({{G}_{k}}-{{\mathcal{G}}_{k}})G_{k}^{-1}(y_{k+v}^{d}-e_{k}^{\mathcal{F}}-{{\mathcal{F}}_{k}}+Cee) \\ 
 & =(I-{{\mathcal{G}}_{k}}G_{k}^{-1})(y_{k+v}^{d}-e_{k}^{\mathcal{F}}-{{\mathcal{F}}_{k}}+\mathcal{C}(e_{k+v-1}^{y})e_{k+v-1}^{y}).
\end{aligned}
\end{equation}
With \eqref{ErrorDynClose} and \eqref{Perturb}, we can derive the error dynamics in control affine form. With some algebraic manipulation on \eqref{Perturb} and \eqref{ErrorDynClose}, we have: 
\begin{equation}
\label{ClosedErrorSystem}
    \left[ \begin{matrix}
   e_{k+v+1}^{y}  \\
   e_{k+1}^{\mathcal{F}}  \\
\end{matrix} \right]={{A}_{k}}\left[ \begin{matrix}
   e_{k+v}^{y}  \\
   e_{k}^{\mathcal{F}}  \\
\end{matrix} \right]+B_k{{R}_{k}},
\end{equation}
where we define ${{H}_{k}}:=\mathcal{G}_{k} G_k^{-1}$ and:
\begin{equation*}
    \begin{matrix}
  {{A}_{k}}:=\left[ \begin{matrix}
   cI-{{H}_{k+1}^{-1}}(I-{{H}_{k}})(1-c) & -{{H}_{k+1}^{-1}}d  \\
   {{H}_{k+1}^{-1}}(I-{{H}_{k}})(1-c) & {{H}_{k+1}^{-1}}d  \\
\end{matrix} \right] \\ 
  B_k=\left[ \begin{matrix}
   I  \\
   -I  \\
\end{matrix} \right],\\
  {{R}_{k}}:={{H}_{k+1}}((I-{{H}_{k}})(y_{k+v}^{d}-{{\mathcal{F}}_{k}}) \\ -(I-{{H}_{k+1}})(y_{k+v+1}^{d}-{{\mathcal{F}}_{k+1}})-\Delta {{F}_{k}}) \\ 
\end{matrix}
\end{equation*}
where $ c:=\mathcal{C}(e_{k+v}^{y}),\ d:=\mathcal{D}(e_{k+v}^{\mathcal{F}})$.
As the input $R_k$ is not a function of the error term, we consider $R_k$ as a perturbation and only analyze the other part for the stability. We  assume here $R_k$ is bounded. Due to the complexity of considering 2 matrix variables, we assume that $H_{k+1} = H_{k} = H$ for simplicity. Therefore we have:
\begin{equation*}
    {{A}_{k}}=\left[ \begin{matrix}
   cI-{{H}^{-1}}(I-H)(1-c) & -d{{H}^{-1}}  \\
   {{H}^{-1}}(I-H)(1-c) & d{{H}^{-1}}  \\
\end{matrix} \right]
\end{equation*}
As we are concerned about the stability at the origin of \eqref{ClosedErrorSystem}, we analyze the linearized dynamics at the origin. Now the problem is to find the matrix $H$ that makes $A_k$ have eigenvalues inside the unit circle. We have the characteristic polynomial: 
\begin{equation}
\label{DetPoly}
\begin{aligned}
  & \det (A-\lambda I)= \\
  &\qquad \det (\left[ \begin{matrix}
   cI-({{H}^{-1}}-I)(1-c)-\Lambda  & -d{{H}^{-1}}  \\
   ({{H}^{-1}}-I)(1-c) & d{{H}^{-1}}-\Lambda   \\
\end{matrix} \right]) \\ 
 & =\det (\left[ \begin{matrix}
   cI-\Lambda  & -\Lambda   \\
   ({{H}^{-1}}-I)(1-c) & d{{H}^{-1}}-\Lambda   \\
\end{matrix} \right]) \\ 
 & =\det (cI-\Lambda )\det (d{{H}^{-1}}-\Lambda \\
 &\qquad \qquad +({{H}^{-1}}-I)(1-c){{(cI-\Lambda )}^{-1}}\Lambda ) \\ 
 & =(c-\lambda )\det (d{{H}^{-1}}-\Lambda +({{H}^{-1}}-I)\frac{(1-c)\lambda }{c-\lambda }) \\ 
 & =\det ({{H}^{-1}}(\frac{(1-c)\lambda }{c-\lambda }+d)-\lambda I-\frac{(1-c)\lambda }{c-\lambda }I), 
\end{aligned}
\end{equation}
where $\Lambda := \lambda I$. As we are concerned about the case when $\|\lambda \| < 1$, then we can assume $I+\Lambda$ is invertible. Additionally, we have $c \in [-1, 1)$, so we have \eqref{DetPoly}. Now we can decompose $H$:
\begin{equation*}
    H^{-1}=P^{-1}JP,
\end{equation*}
where $J$ is the Jordan canonical form and $P$ is an invertible matrix. Therefore we have:
\begin{equation*}
\begin{aligned}
  & {{H}^{-1}}(\frac{(1-c)\lambda }{c-\lambda }+d)-\lambda I-\frac{(1-c)\lambda }{c-\lambda }I \\ 
 & ={{P}^{-1}}\left( J(\frac{(1-c)\lambda }{c-\lambda }+d)-(\lambda +\frac{(1-c)\lambda }{c-\lambda })I \right)P  
\end{aligned}
\end{equation*}
Let the eigenvalues of $H$ be $\alpha_{j} \in \mathbb{C}, j = 1,2,...,n$, then $\det (A-\lambda I) = 0$ is equivalent to: 
\begin{equation}
\label{eigenSol}
    \prod_{j=1}^{n} \left(\alpha_j {{\lambda }^{2}}+(1-c-d-\alpha_j )\lambda +dc \right)= 0,\ \ \
    \|\lambda\|<1.
\end{equation}
When $c=d=-1$, if each $\alpha_j$ ensures solutions to the quadratic function $\alpha_j {{\lambda }^{2}}+(1-c-d-\alpha_j )\lambda +dc =0$ being inside of unit circle, the error dynamics \eqref{ClosedErrorSystem} would be at least locally asymptotically stable. 
We notice that each $\alpha_j$ corresponds to a pair of solution $\lambda$. For simplicity below, we omit the subscript $j$ and analyze each $\alpha$. 



\section{Numerical Analysis}
\label{simresult}

\begin{figure}[t]
    \centering
    \includegraphics[width=1\columnwidth]{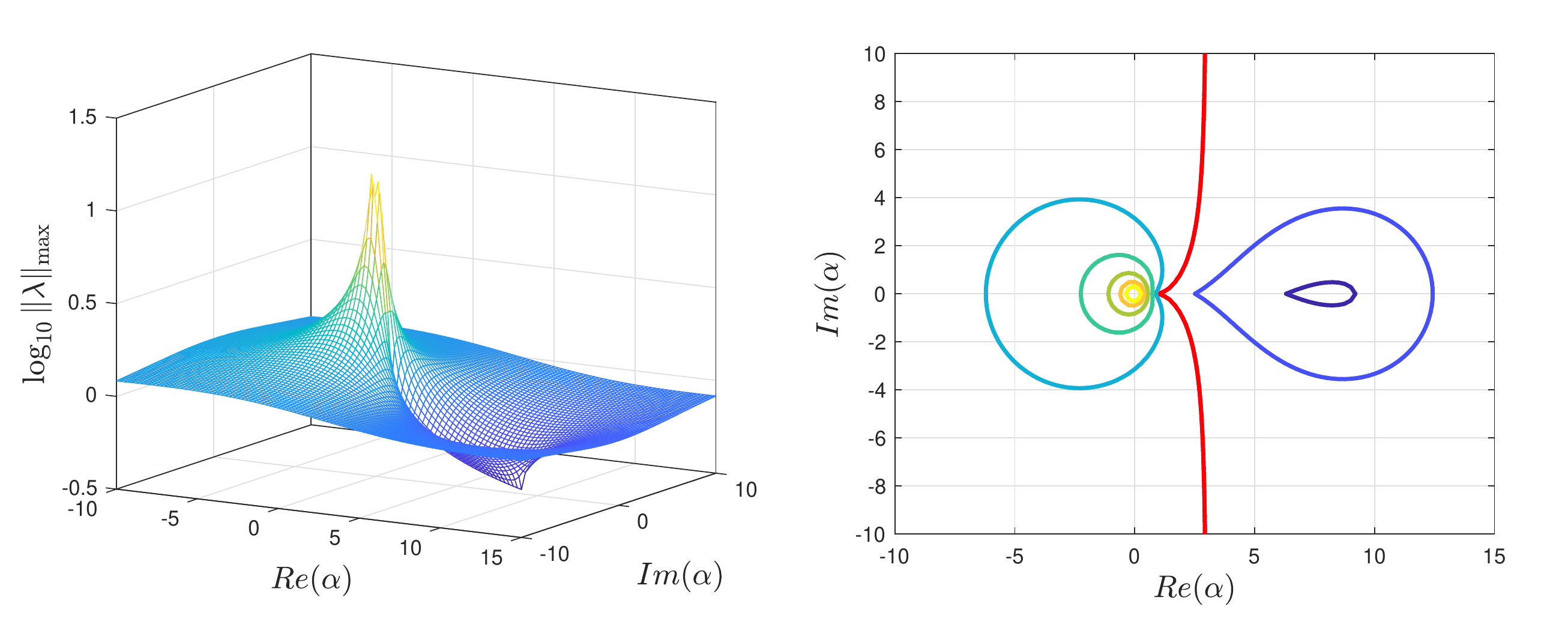}
    \caption{Norm of eigenvalues of $A_k$, i.e. $\lambda$ w.r.t the eigenvalues of $H$, i.e. $\alpha$ when $c=d=-1$. Left: The distribution of $\|\lambda\|_{\max}$ in complex plane. Right: Contour of distribution of $\|\lambda\|_{\max}$ in complex plane. $\alpha$ at the right side of the red contour ensure the error dynamics \eqref{ErrorDynClose} is stable at the origin.}
    \label{fig:eigenFullH}
\end{figure}

Now we numerically analyze the effect of $\mathcal{G}_k$ on the stability and launch several simulations to verify it.
\subsection{Pole distribution}
To make \eqref{eigenSol} more intuitive, we sample $\alpha$ and plot the corresponding $\|\lambda\|$ in Fig.~\ref{fig:eigenFullH}. Note that each $\alpha$ corresponds to 2 (distinct) solutions of $\lambda$ in \eqref{eigenSol}: 
\begin{equation}
    \lambda =\frac{(\alpha +d+c-1)\pm \sqrt{{{(\alpha +d+c-1 )}^{2}}-4dc\alpha  }}{2\alpha }.
\end{equation}
Thus we only consider the larger norm, namely $\|\lambda\|_{\max}$. For the stability at the origin, we let $c=d=-1$ and we can see $\alpha$ at the right side of the red contour in Fig.~\ref{fig:eigenFullH} can ensure the all $\lambda$ being inside of the unit circle. The boundary (red contour) can also be obtained via
    $\alpha =\frac{(1-c-d)\lambda +dc}{\lambda -{{\lambda }^{2}}}$,
when we make $\|\lambda\| = 1$ in the complex plane. 

\begin{figure}[t]
    \centering
    \includegraphics[width=0.6\columnwidth]{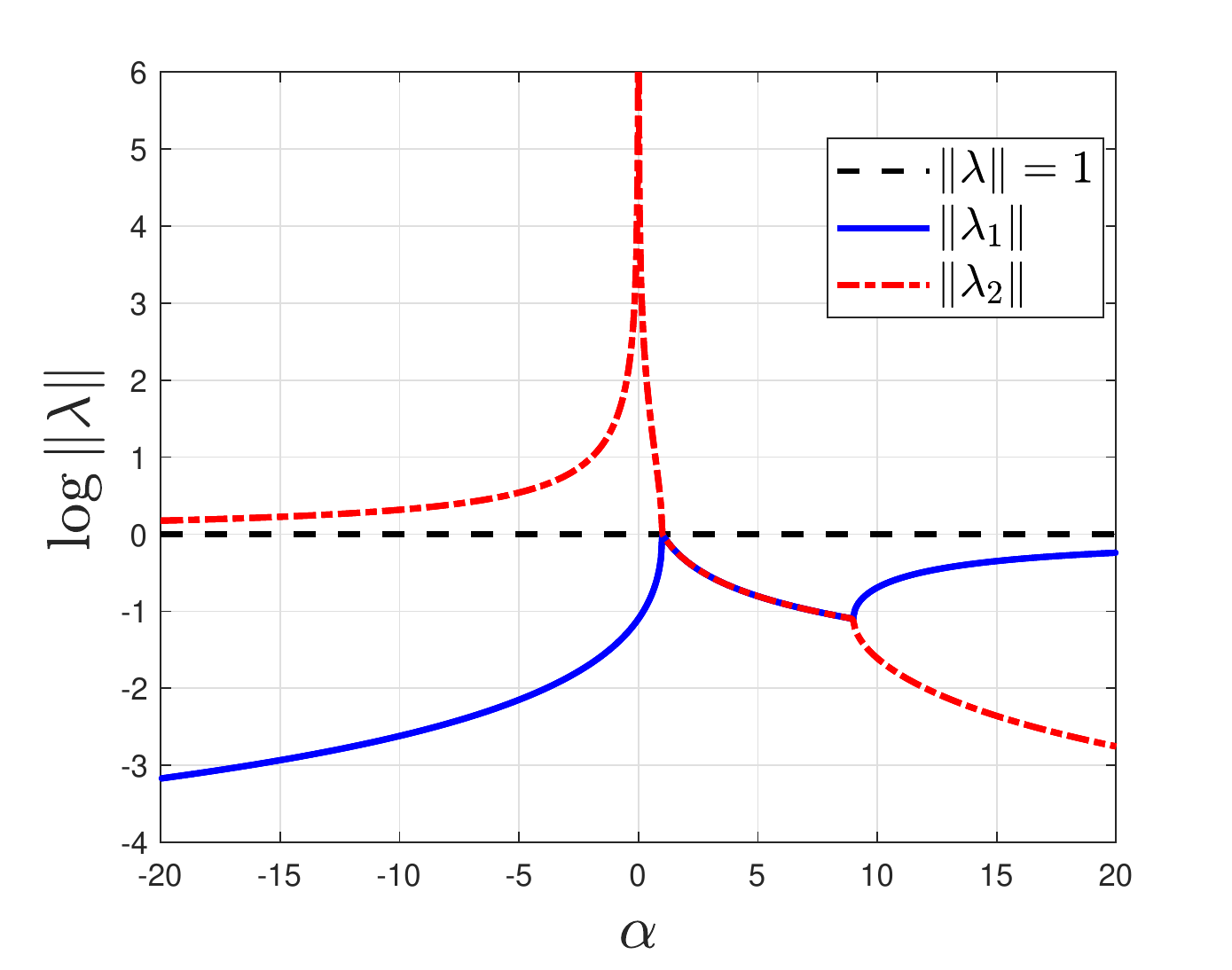}
    \caption{Norm of eigenvalues of $A_k$ when $\alpha \in \mathbb{R}$ and $c=d=-1$. }
    \label{fig:eigs_origin}
\end{figure}
\begin{figure*}
    \centering
    \includegraphics[width=1.95\columnwidth]{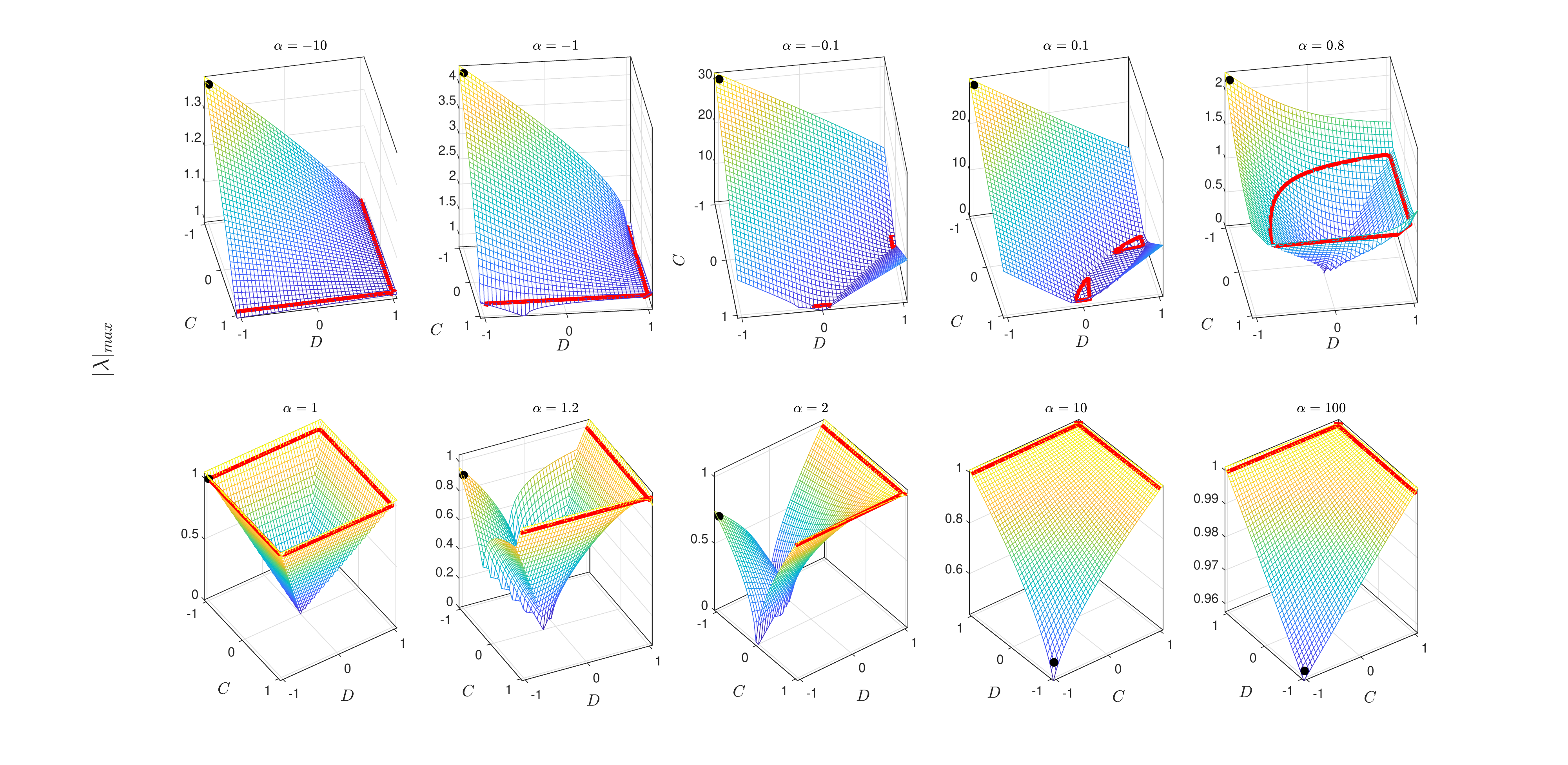} 
    \caption{The distribution of $\|\lambda\|_{\max}$ in the state space of system \eqref{ClosedErrorSystem}. Red contour: $\|\lambda\|_{\max} = 1$. Lighter region: larger $\|\lambda\|_{\max}$. Darker region: smaller $\|\lambda\|_{\max}$. Black dots: Origin of the system \eqref{ClosedErrorSystem}: $c = d = -1$. When $\alpha > 1$, $\|\lambda\|_{\max} < 1$ in the entire space except for $d = c = 1$. However, when $\alpha \rightarrow +\infty$, $\|\lambda\|_{\max} \rightarrow 1$, thus the system tends to be unstable. When $\alpha < 1$, the origin is no longer attractive. However, $\|\lambda\|_{\max} < 1$ still holds for some part of the state space (the enclosed region by red contours in Fig. \ref{fig:eigs}) thus the error may not go to infinity. As $\alpha \rightarrow -\infty$, this region vanishes.}
    \label{fig:eigs}
    \vspace{-4mm}
\end{figure*}

We also present the case that $Im(\alpha) = 0$ in Fig. \ref{fig:eigs_origin}. We can see that the origin is stable when $\alpha > 1$. This result suggests that if we only overestimate the scale of $G_k$ and apply it for control, the origin can be stable. But if $\alpha \rightarrow +\infty$, $\|\lambda_1\|$ will approach 1, which makes the origin sensitive to disturbance. When $\alpha < 1$, $\|\lambda_1\| > 1$, the origin becomes unstable. 

Note that the $A_k$ is a function of the error term, which makes constructing the Lyapunov function for \eqref{ClosedErrorSystem} difficult. We instead analyze the eigenvalues of $A_k$ to roughly show the global properties of \eqref{ClosedErrorSystem}. We fix $\alpha \in \mathbb{R}$ and plot $\|\lambda\|_{\max}$ with respect to $c$ and $d$ in Fig. \ref{fig:eigs}. As $c$ and $d$ are monotone functions of $\|e^{y}_{k+v}\|$ and $\|e^{\mathcal{F}}_k\|$, Fig. \ref{fig:eigs} can reflect the distribution of $\|\lambda_{\max}\|$ in the entire space of system \eqref{ClosedErrorSystem}. 

\begin{figure}[t]
    \centering
    \includegraphics[width=0.45\columnwidth]{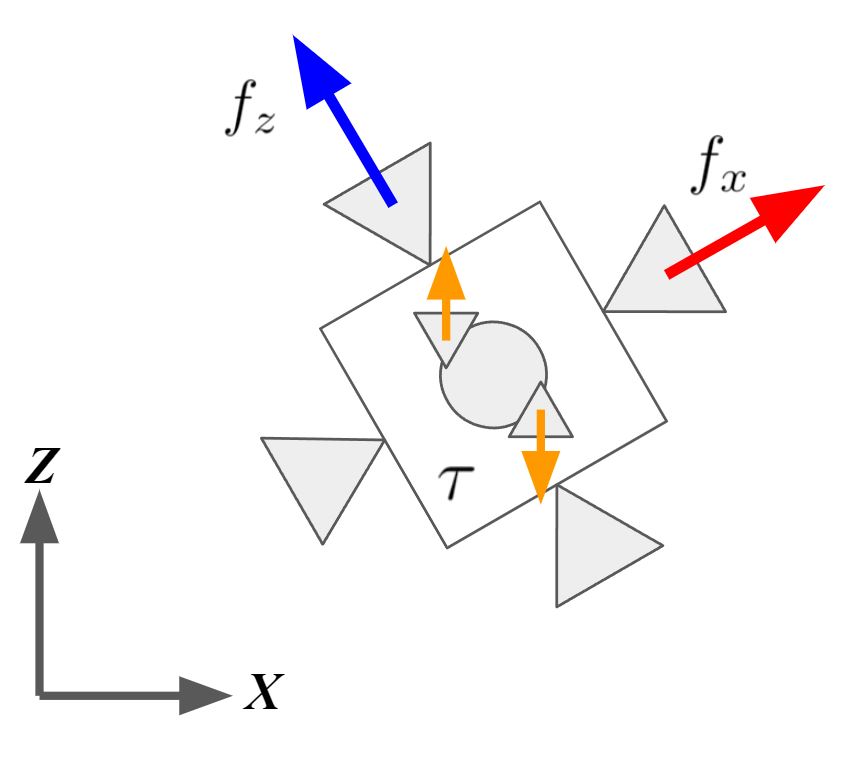}
    \caption{2D fully actuated rigid body system. The system inputs are force $(f_x, f_z)$ and torque $\tau$ applied in the body frame. For the purpose of simulation, we let the mass be $2.0\ kg$ and the inertial be $3.0\ kg\ m^2$.}
    \label{fig:rigid}
    \vspace{-3mm}
\end{figure}


\begin{figure}[t]
    \centering
    \includegraphics[width=0.9\columnwidth]{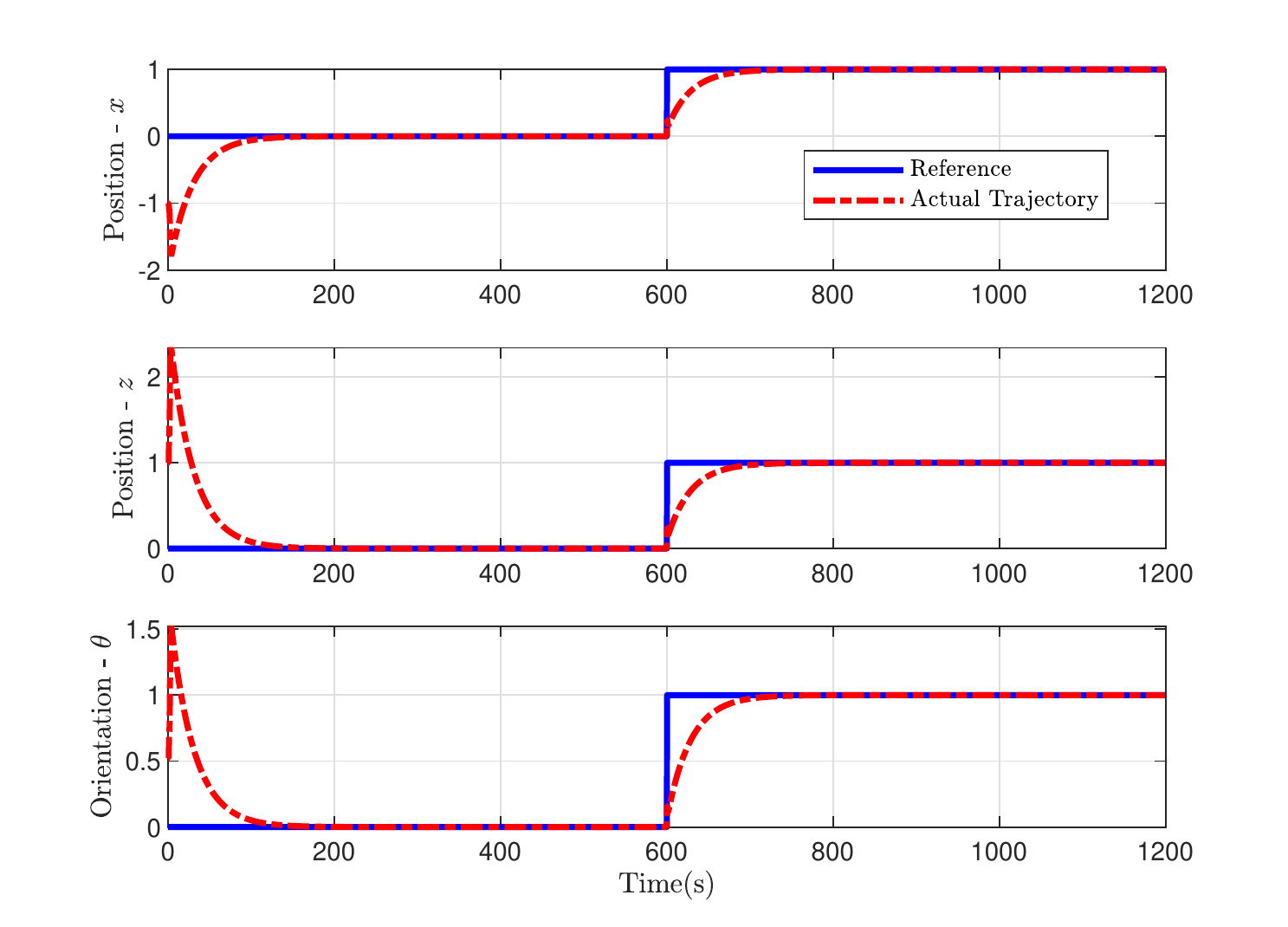}
    \caption{Reference step signal, initial condition and tracking performance when $\alpha = 1.2$. }
    \label{fig:reftraj}
    \vspace{-3mm}
\end{figure}

\subsection{2D fully actuated rigid body}

To validate the former analysis, we apply the ULM-FTS framework with different $\alpha \in \mathbb{R}$ to an input-output controllable system. We consider a 2D rigid body system, where the states are the Cartesian position $(x, z)$ and orientation $\theta$:
\begin{equation*}
\left[\begin{array}{l}
\ddot{x} \\
\ddot{z} \\
\ddot{\theta}
\end{array}\right]=\left[\begin{array}{ccc}
\frac{1}{m} \cos \theta & \frac{1}{m} \sin \theta & 0 \\
-\frac{1}{m} \sin \theta & \frac{1}{m} \cos \theta & 0 \\
0 & 0 & \frac{1}{I}
\end{array}\right]\left[\begin{array}{l}
f_{x} \\
f_{z} \\
\tau
\end{array}\right].
\end{equation*}The system is presented in Fig. \ref{fig:rigid}.
\begin{figure*}
    \centering
    \includegraphics[width=1.9\columnwidth]{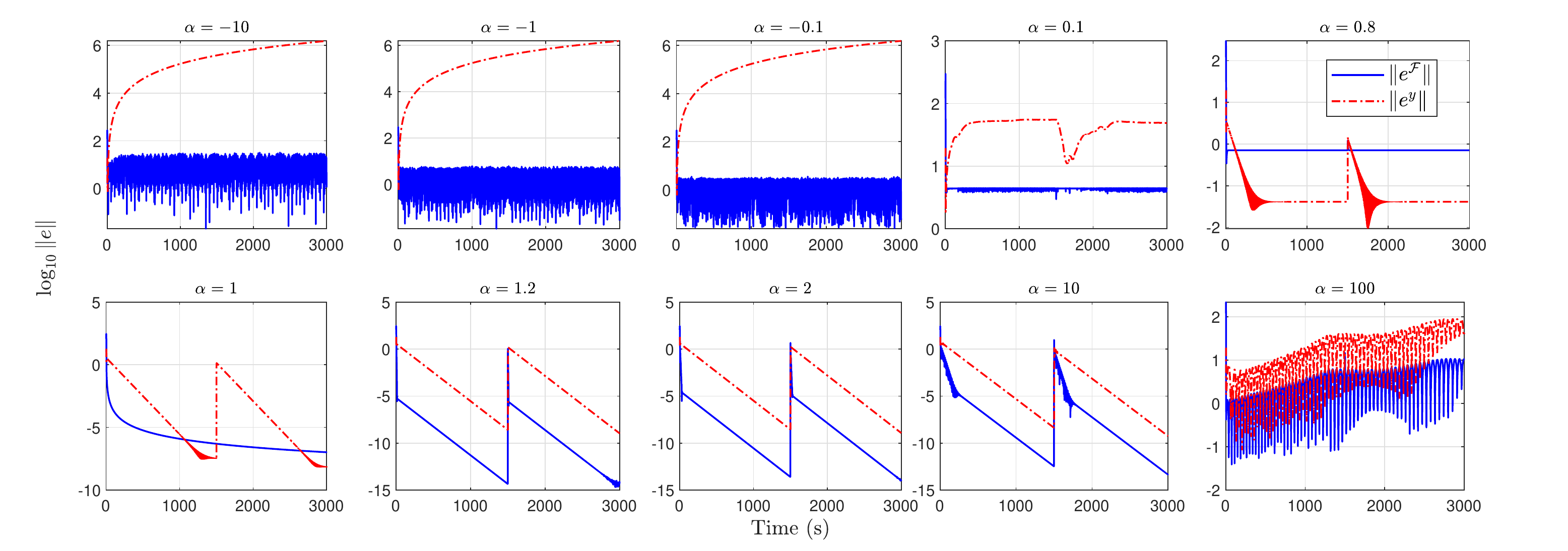}
    \caption{Closed-loop simulation with different $\alpha \in \mathbb{R}$. When $\alpha < 1$, the error do not converge to 0 and may diverge completely. When $\alpha = 1$, the perturbation term $R_k$ becomes zero so the state estimation error consistently converge to $0$. Note that the jump in tracking error is solely caused by the saturation of inputs. When $\alpha = 1.2, 2, 10$, the error is asymptotically stable according to $\|\lambda\|_{\max}$ at the origin. When $\alpha = 100$, $\|\lambda\|_{\max}$ at the origin is near 1, which makes system \eqref{ClosedErrorSystem} extremely sensitive to $R_k$. }
    \label{fig:errorsim}
    \vspace{-3mm}
\end{figure*}
We define the time step as $\Delta t:=t_{k+1} - t_{k}$ and we get the following discrete dynamical system:
\begin{equation*}
\begin{aligned}
  {{y}_{k}^{(2)}}&=\frac{{{y}_{k+2}}-2{{y}_{k+1}}+{{y}_{k}}}{\Delta {{t}^{2}}} \\ 
 & =\left[ \begin{matrix}
   \tfrac{1}{m}\cos {{\theta }_{k}} & \tfrac{1}{m}\sin {{\theta }_{k}} & 0  \\
   -\tfrac{1}{m}\sin {{\theta }_{k}} & \tfrac{1}{m}\cos {{\theta }_{k}} & 0  \\
   0 & 0 & \tfrac{1}{I}  \\
\end{matrix} \right]\left[ \begin{matrix}
   {{f}_{x,k}}  \\
   {{f}_{z,k}}  \\
   {{\tau }_{k}}  \\
\end{matrix} \right] .
\end{aligned}
\end{equation*}
Converting to the form in \eqref{eq:system}, we have
\begin{equation*}
\begin{aligned}
  & {{y}_{k+2}}={{F}_{k}}+{{G}_{k}}{{u}_{k}}, \\ 
 & {{G}_{k}}=\Delta {{t}^{2}}\left[ \begin{matrix}
   \tfrac{1}{m}\cos {{\theta }_{k}} & \tfrac{1}{m}\sin {{\theta }_{k}} & 0  \\
   -\tfrac{1}{m}\sin {{\theta }_{k}} & \tfrac{1}{m}\cos {{\theta }_{k}} & 0  \\
   0 & 0 & \tfrac{1}{I}  \\
\end{matrix} \right] ,\\ 
 & {{F}_{k}}=2{{y}_{k+1}}-{{y}_{k}},\ \ {{u}_{k}}={{\left[ \begin{matrix}
   {{f}_{x,k}} & {{f}_{z,k}} & {{\tau }_{k}}  \\
\end{matrix} \right]}^{T}} .\\ 
\end{aligned}
\end{equation*}
We apply the controller with different $\alpha$, i.e. $\mathcal{G}_k = \alpha G_k$. Two step signals are designed for the controller to track, see Fig. \ref{fig:reftraj}. We also assume upper bounds of the input, i.e., $\|u_k\|_{\infty} \leq 3$.

We apply the controller with different input influence matrices parameterized by $\alpha \in \mathbb{R}$. The tracking errors are presented in Fig.~\ref{fig:errorsim}. The result is consistent with the distribution of eigenvalues of $A_k$, shown in Fig.~\ref{fig:eigs_origin} and Fig.~\ref{fig:eigs}. When $\alpha = 1.2, 2, 10$, the tracking error soon converges as $\|\lambda\|_{\max} < 1$. It is worth noticing that $\alpha = 1$ means that the $R_k = 0$. Therefore, $\|e^{\mathcal{F}}_k\|$ is continuously dropping. The jump in the tracking error is due to the saturation of input in this case. When $\alpha < 1$, the origin is no longer stable, and thus the tracking error is much larger or diverges. As $\alpha \rightarrow -\infty$, the tracking error starts to diverge. Note that when $\alpha = 100$, the $A_k$ have poles with the norm approaching 1 at the origin; thus, the system is sensitive to perturbations.




\section{Discussion}
\label{sec:discuss}
The previous result suggests that if the designed input influence matrix is within some range w.r.t the real input influence matrix, we can guarantee asymptotically stable error dynamics. One practical application is when partial knowledge of the system is given. Based on Fig.~\ref{fig:eigenFullH}, we can overestimate the scale of the input influence matrix for control. For example, multiply the imperfectly known influence matrix with a positive number to make the eigenvalues of $H$ reside in the valley in the right half-plane of Fig.~\ref{fig:eigenFullH}.

This work assumes the use of a discrete-time Hölder-continuous finite-time stable control and estimation scheme; it can be extended to other ULM-based model-free control and state estimation cases. One example is that we replace the gain in \eqref{FTScontroller} and \eqref{FTSObserver} with constants between $-1$ and $1$. In this case, the error dynamics \eqref{ClosedErrorSystem} becomes linear and can be globally asymptotically stable. 

As it is hard to construct the Lyapunov function for the error dynamics \eqref{ClosedErrorSystem}, we only gives the local stability by the eigenvalues of $A_k$ matrix. As the $A_k$ matrix in \eqref{ClosedErrorSystem} remains Hurwitz for a wide range of $c$ and $d$ given $H$ (see Fig.~\ref{fig:eigs}), we may extend the local asymptotic stability to global. 

%

%
One limitation of this work is that the analysis only applies to control affine systems, where we could define the actual system input influence matrix and dynamics. For system that is not control affine, we have to compare the ULM dynamics with the actual system without this assumption for a more generalized stability criterion. Another limitation of this work is the assumption that $H_{k+1} = H_{k}$ in \eqref{ClosedErrorSystem}. Without this simplification, $A_k$ in \eqref{ClosedErrorSystem} will explicitly depend on time. {However, if $A_k$ is Lipschitz continuous or weakly dependent on time, we may be able to obtain the stability criteria using the same framework.} All these problems indicate interesting future research directions. 

\section{Conclusion}
\label{sec:conclusion}
This paper analyzes the stability of ultra-local model-based model-free control for discrete-time control affine MIMO system by extracting the input-dependent part from the ULM dynamics. In the case of applying the H\"older-continuous finite-time stable controller, we find that the error convergence is determined by the eigenvalues of the difference (defined by matrix multiplication) between the designed and real input influence matrix. We show that this condition is not conservative. This result can guide designing the input influence matrix for MIMO ULM-based model-free control when only partial knowledge of the system is accessible.

{\footnotesize 
\bibliographystyle{IEEEtran}
\bibliography{bib/strings-abrv,bib/ieee-abrv,bib/references}

\begin{thebibliography}{10}
\providecommand{\url}[1]{#1}
\csname url@rmstyle\endcsname
\providecommand{\newblock}{\relax}
\providecommand{\bibinfo}[2]{#2}
\providecommand\BIBentrySTDinterwordspacing{\spaceskip=0pt\relax}
\providecommand\BIBentryALTinterwordstretchfactor{4}
\providecommand\BIBentryALTinterwordspacing{\spaceskip=\fontdimen2\font plus
\BIBentryALTinterwordstretchfactor\fontdimen3\font minus
  \fontdimen4\font\relax}
\providecommand\BIBforeignlanguage[2]{{%
\expandafter\ifx\csname l@#1\endcsname\relax
\typeout{** WARNING: IEEEtran.bst: No hyphenation pattern has been}%
\typeout{** loaded for the language `#1'. Using the pattern for}%
\typeout{** the default language instead.}%
\else
\language=\csname l@#1\endcsname
\fi
#2}}

\bibitem{sanyal2021discrete}
A.~K. Sanyal, ``Discrete-time data-driven control with {H{\"o}lder}-continuous
  real-time learning,'' \emph{Int. Journal of Control}, pp. 1--13, 2021.

\bibitem{ulm}
M.~Fliess and C.~Join, ``Model-free control,'' \emph{Int. Journal of Control},
  vol.~86, no.~12, pp. 2228--2252, 2013.

\bibitem{iPID}
M.~Fliess, ``Model-free control and intelligent pid controllers: towards a
  possible trivialization of nonlinear control?'' \emph{IFAC Proceedings
  Volumes}, vol.~42, no.~10, pp. 1531--1550, 2009.

\bibitem{ULM-siso1}
C.-H. Chang, V.~H. Duenas, and A.~K. Sanyal, ``Model free nonlinear control
  with finite-time estimation applied to closed-loop electrical stimulation
  induced cycling,'' in \emph{Proc. Amer. Control Conf.}, 2020, pp. 5182--5187.

\bibitem{ulm-siso2}
A.~N. Chand, M.~Kawanishi, and T.~Narikiyo, ``Non-linear model-free control of
  flapping wing flying robot using {iPID},'' in \emph{Proc. {IEEE} Int. Conf.
  Robot. and Automation}.\hskip 1em plus 0.5em minus 0.4em\relax IEEE, 2016,
  pp. 2930--2937.

\bibitem{ULM-MPC}
Z.~Wang and J.~Wang, ``Ultra-local model predictive control: A model-free
  approach and its application on automated vehicle trajectory tracking,''
  \emph{Control Engineering Practice}, vol. 101, p. 104482, 2020.

\bibitem{ULM-LMI}
Y.~Al~Younes, A.~Rabhi, and H.~Al-Wedyan, ``Intelligent controller design for
  {MIMO} systems using model-free control and {LMI} approaches applied on a
  twin rotor {MIMO} system,'' in \emph{Advances in Science and Engineering
  Technology Int. Conf.}\hskip 1em plus 0.5em minus 0.4em\relax IEEE, 2019, pp.
  1--6.

\bibitem{fts-lyap}
S.~P. Bhat and D.~S. Bernstein, ``Finite-time stability of continuous
  autonomous systems,'' \emph{SIAM Journal on Control and Optimization},
  vol.~38, no.~3, pp. 751--766, 2000.

\bibitem{cfts1}
A.~K. Sanyal and J.~Bohn, ``Finite-time stabilisation of simple mechanical
  systems using continuous feedback,'' \emph{Int. Journal of Control}, vol.~88,
  no.~4, pp. 783--791, 2015.

\bibitem{fts2}
S.~P. Viswanathan, A.~K. Sanyal, and R.~R. Warier, ``Finite-time stable
  tracking control for a class of underactuated aerial vehicles in {SE(3)},''
  in \emph{Proc. Amer. Control Conf.}\hskip 1em plus 0.5em minus 0.4em\relax
  IEEE, 2017, pp. 3926--3931.

\bibitem{fts3}
R.~Hamrah, A.~K. Sanyal, and S.~P. Viswanathan, ``Discrete finite-time stable
  attitude tracking control of unmanned vehicles on {SO(3)},'' in \emph{Proc.
  Amer. Control Conf.}\hskip 1em plus 0.5em minus 0.4em\relax IEEE, 2020, pp.
  824--829.

\bibitem{fts-obs}
N.~Wang and A.~K. Sanyal, ``A {H{\"o}lder}-continuous extended state observer
  for model-free position tracking control,'' in \emph{Proc. Amer. Control
  Conf.}\hskip 1em plus 0.5em minus 0.4em\relax IEEE, 2021, pp. 2133--2138.

\end{thebibliography}
}

\end{document}